\documentclass[letterpaper, 10 pt, conference]{ieeeconf}  

\IEEEoverridecommandlockouts                              

\usepackage{graphics} 
\usepackage{epsfig} 
\usepackage{times} 
\usepackage[ruled,vlined]{algorithm2e}
\usepackage{amsmath} 
\usepackage{amssymb}  
\usepackage{hyperref}
\usepackage{tabularx}
\usepackage{xcolor}
\usepackage{array}
\newtheorem{definition}{Definition}
\newtheorem{example}{Example}

\newtheorem{problem}{Problem}

\usepackage{booktabs} 
\usepackage{subcaption}
\usepackage{multirow}
\usepackage[subtle,tracking=normal]{savetrees}

\usepackage[normalem]{ulem}

\newcommand\AP{\mathcal{P}}

\usepackage[acronym]{glossaries}
\makeglossaries
\newacronym{WPMSAT}{WPMSAT}{Partial Weighted MaxSAT}
\title{\LARGE \bf
Learning Linear Temporal Specifications \\ from Demonstrations with Uncertainty
}

\author{Parastou Fahim$^{1}$, Constantino Lagoa$^{1}$, and R\^{o}mulo Meira-G\'{o}es$^{1}$
\thanks{*This research was supported by Rising Researcher award ICDS:RR25-SCR025154 from Penn State's Institute for Computational \& Data Sciences.}
\thanks{$^{1}$PF, CL, and RMG are with the School of Electrical Engineering and Computer Science at
The Pennsylvania State University, State College, USA
{\tt\small \{pbf5107,cml18,romulo\}@psu.edu}}%
}

\begin{document}

\maketitle
\thispagestyle{empty}
\pagestyle{empty}

\begin{abstract}
Learning temporal logic specifications from system demonstrations is essential for tasks such as formal verification and controller synthesis, especially in safety-critical domains. 
Existing approaches typically assume demonstrations are correct or only affected by misclassification errors. In practice, however, system traces are often uncertain or incomplete due to sensor faults, measurement errors, or data loss.
We present a framework for learning minimal Linear Temporal Logic (LTL) formulas from demonstrations with uncertainty. 
Our approach models uncertainty via Hamming distance to generate possible estimates around each observed trace, which are grouped with constraints requiring that at least one trace per group is consistent with the learned formula. 
Our problem is then reduced to an equivalent Pseudo-Boolean Optimization.
We evaluate our method against state-of-the-art LTL learning approaches and show that it recovers specifications that more closely align with ground-truth formulas under uncertainty.
\end{abstract}

\section{INTRODUCTION}

The challenge of inferring temporal logic specifications from system observations has gained increasing attention in recent years, particularly in areas such as robotics, control systems, and cyber-physical security  \cite{puranic2021learning, bartocci2022survey, bordais2024complexity}. 
Learning an interpretable specification from system behavior is essential for facilitating formal verification and controller synthesis \cite{belta2021formal}.
 This need is heightened in safety-critical applications, such as autonomous vehicles, unmanned aerial systems, and medical devices, where incorrect or unforeseen behaviors can result in catastrophic outcomes \cite{wolff2014control}.
As system complexity grows, manually specifying temporal behaviors becomes challenging and error-prone. 
Automatically inferring temporal logic specifications from system executions provides a scalable alternative. This enables constructing human-interpretable models of system behaviors, the identification of deviations or faults, and the formal verification of system correctness \cite{neider2018learning, bordais2023learning}.

Linear temporal logic (LTL) has emerged as a specification formalism for capturing system behaviors over time. 
Recent studies have demonstrated that LTL properties can be automatically learned from observed executions, typically using both positive and negative examples \cite{neider2018learning,gaglione2022maxsat,ghiorzi2023learning}. 
Two common frameworks for the learning task are \textit{automata-based learning} and \textit{SAT-based learning}. 
Automata-based methods operate by translating LTL formulas into automata whose accepted languages correspond to the set of traces that satisfy the given specifications. 
Depending on the complexity of the temporal behaviors to be captured, different types of automata may be employed,  such as finite state automata (FSA) \cite{bhatia2010sampling}, Büchi automata \cite{tsay2022linear}, or Rabin automata \cite{azzopardi2024direct}, each defined by distinct acceptance conditions tailored to varying levels of expressivity \cite{belta2021formal}.
In this paper, we will focus on SAT-based learning methods since they usually offer a more direct and scalable approach to learn LTL formulas. 






SAT-based learning methods encode the task of learning an LTL formula as a satisfiability problem (SAT) \cite{neider2018learning}. 
Many recent studies have adopted SAT-based learning methods because they provide a practical balance between computational performance and model explainability, particularly in large-scale scenarios \cite{shi2021transformer}. 
For example, the study by Camacho and McIlraith \cite{camacho2019learning} proposes a SAT-based framework for both passive and active learning of minimal $\mathrm{LTL}_f$ formulas from positive and negative examples. 
Their approach utilizes alternating finite automata (AFA) to construct minimal and interpretable models that generalize from limited data.
The work in \cite{ghiorzi2023learning} focuses on improving the scalability of learning LTL specifications, compared to \cite{neider2018learning,gaglione2022maxsat}.
Applied to robotic task failure analysis, the method demonstrates practical effectiveness and achieves faster performance while maintaining accuracy.

To ensure that learned temporal specifications reflect real-world behavior, it is important to account for uncertainty in the input data. 
Existing research has proposed several approaches to address issues such as label noise or dummy variables. 
Neider and Gavran \cite{neider2018learning} address this challenge through two approaches: a SAT-based method that finds minimal consistent LTL formulas, and a decision tree-based method that improves scalability by combining learned LTL primitives. 
To evaluate robustness, they introduce noise by injecting an unconstrained atomic proposition into traces. 
This simulates irrelevant or spurious signals, and their methods successfully infer correct formulas despite this perturbation.
The work in \cite{gaglione2022maxsat} takes a different direction by framing the LTL and STL specification learning problem as a MaxSAT/MaxSMT optimization task.
Their approach supports learning formulas under misclassification error, e.g., a positive trace that is incorrectly labeled negative.
Related work has also explored learning temporal logic and structured behavioral models from demonstrations with different assumptions, including demonstrations that satisfy the specification but may be cost-suboptimal, for example, in path length \cite{chou2022learning}; continuous state/action trajectories rather than discrete symbolic traces \cite{li2021learning}; and, partial observability, where the agent has incomplete state information \cite{toro2019learning}.
 

However, previous works primarily focus on misclassification error, assuming that the traces themselves remain intact and correctly captured.
These methods do not address a common real-world challenge: the presence of corrupted, incomplete, or inconsistent traces caused by sensor faults, data loss, or measurement errors, i.e., uncertainty in the demonstration traces. 
This highlights the need for models that can learn temporal logic specifications even when trace content is uncertain, not just their classification, capturing a more realistic form of robustness for practical applications.




In this paper, we start to address the problem of learning minimal LTL specifications from demonstrations with uncertainty.
First, we model trace uncertainty using the Hamming distance to capture the distance between an observed demonstration and the possible estimates that can generate the observation.
The combination of observed trace and the possible estimates within a maximum distance bound forms an \emph{uncertainty group}.
Next, we integrate these uncertainty groups into the LTL learning problem from traces with uncertainty, which we call the robust LTL learning problem.
Unlike prior work that assumes every observed trace perfectly reflects the true system behavior, or only models misclassification errors, our formulation enforces that at least one trace within each uncertainty group must satisfy the learned LTL formula.

To solve our robust LTL learning problem, we transform it to an equivalent Pseudo-Boolean Optimization, for which we obtain a solution.
We implement our solution methodology and compare it to the state-of-the-art SAT-learning methods.
Our results show that, unlike existing approaches, our method can learn LTL specifications from uncertain traces that more closely match the ground-truth formulas.
In this way, our framework bridges the gap between theoretical LTL learning and practical scenarios by effectively capturing uncertainty during specification learning.


The remainder of this paper is organized as follows: Section~\ref{sec:preliminaries} introduces essential preliminaries on LTL and related mathematical concepts. 
Section~\ref{sec:problem} formally defines the robust LTL learning problem considered in this work. 
Section~\ref{sec:solution} details our proposed learning approach, describing how we incorporate trace uncertainties into the learning process. 
In Section~\ref{sec:experiments}, we evaluate our method against a state-of-the-art LTL learning method.
Finally, we conclude our paper in Section~\ref{sec:conclusion} with directions for future research.

\section{Preliminaries}
\label{sec:preliminaries}
In this section, we establish the necessary background concepts and notation that form the basis of our work.


\vspace{-0.45em}
\subsection{Propositional Boolean Logic}

    Let  $\mathcal{V}$ be a set of propositional variables that take Boolean values from \( \mathbb{B} = \{0, 1\} \), where 0 represents \emph{false} and 1 represents \emph{true}. Formulas in \emph{propositional logic} are inductively constructed as follows:

\begin{itemize}
    \item Each  $x \in \mathcal{V}$
 is a propositional formula.
    \item If \( \Psi \) and \( \Phi \) are propositional formulas, then so are \( \neg \Psi \) (negation) and \( \Psi \lor \Phi \) (disjunction).
\end{itemize}

Additionally, we allow the usual Boolean connectives such as \( \Psi \land \Phi \) (conjunction), \( \Psi \Rightarrow \Phi \) (implication), and \( \Psi \Leftrightarrow \Phi \) (equivalence).

\subsection{Propositional Valuation and Satisfiability}

A \emph{propositional valuation} is a mapping $V : \mathcal{V} \to \mathbb{B}$
 , which assigns a Boolean value to each propositional variable. The semantics of propositional logic are defined through the \emph{satisfaction relation} \( \models \) as follows:

\begin{itemize}
    \item \( V \models x \) if and only if \( V(x) = 1 \).
    \item \( V \models \neg \Psi \) if and only if \( V \not\models \Psi \).
    \item \( V \models \Psi \lor \Phi \) if and only if \( V \models \Psi \) or \( V \models \Phi \).
\end{itemize}

A propositional formula \( \Phi \) is \emph{satisfiable} if there exists a valuation \( V \) such that \( V \models \Phi \). In this case, we say that \( V \) is a \emph{model} of \( \Phi \). The \emph{size} of a formula is defined as the number of its subformulas.
\vspace{-0.45em}
\subsection{Finite Traces}
A \emph{trace} represents a sequence of observations of a system.
Formally, a trace is a finite sequence of propositional variables $\mathcal{P}$, i.e., $tr = a_0\dots a_n$ where $a_i\in 2^{\mathcal{P}}$.
The empty trace $\epsilon$ represents an empty sequence, while the length of a trace is given by $|tr|$.
The set of all possible traces is denoted by $(2^{\AP})^*$ where as $(2^{\AP})^n$ denotes all possible traces of length $n+1\in \mathbb{N}$.
For a trace $tr \in (2^{\AP})^*$, $tr[i]$ denotes the $i^{th}$ symbol position, i.e., $tr = a_0\dots a_n$ then $tr[3] = a_3$.
\vspace{-0.45em}
\subsection{Linear Temporal Logic}
While propositional logic offers a static view of Boolean valuations, real-world systems inherently evolve through time. 
For this reason, propositional logic is augmented with temporal operators, forming Linear Temporal Logic (LTL), which enables reasoning about the evolution of propositions (traces) across discrete time steps \cite{BaierKatoen2008}.

\noindent
\textit{Syntax of LTL.}
LTL formulas are constructed according to the following grammar and propositional variables $\AP$ \cite{BaierKatoen2008}:
\[
\varphi ::= p \mid \neg \varphi \mid \varphi \lor \psi \mid \mathbf{X} \varphi \mid \varphi \mathbf{U} \psi, \quad \text{where } p \in \mathcal{P}.
\]

Any propositional variable $p \in \mathcal{P}$ is itself an LTL formula. 
If $\varphi$ and $\psi$ are LTL formulas, then $\neg \varphi$, $\varphi \lor \psi$, $\mathbf{X}\varphi$ ("next"), and $\varphi \mathbf{U} \psi$ ("until") are also LTL formulas.
Other useful temporal operators can be defined based on these fundamental ones, notably:
Eventually -- $\mathbf{F}\varphi$, expressing that formula $\varphi$ must become true at some future point.
Globally -- $\mathbf{G}\varphi$, stating that $\varphi$ must remain true in every future state.

\noindent
\textit{Semantics of LTL.}
Herein, similarly to \cite{gaglione2022maxsat}, we use the finite trace semantics of LTL from \cite{Giacomo:2013}.

The satisfaction of an LTL formula by a trace at position $i$, denoted $(tr,i) \models \varphi$, is defined inductively:
\begin{itemize}
\item $(tr,i) \models p$ if and only if $p \in tr[i]$.
\item $(tr,i) \models \neg \varphi$ if and only if $(tr,i) \not\models \varphi$.
\item $(tr,i) \models \varphi \lor \psi$ if and only if $(tr,i) \models \varphi$ or $(tr,i) \models \psi$.
\item $(tr,i) \models \mathbf{X}\varphi$ if and only if $i < |tr|$ and $(tr,i+1) \models \varphi$.
\item $(tr,i) \models \varphi \mathbf{U}\psi$ if and only if $\exists j$ such that $i\leq j\leq |tr|$ we have $(tr,j) \models \psi$ and $\forall k$ in $i \le k < j$, $(tr,k) \models \varphi$.
\end{itemize}

We say that a trace $tr$ satisfies an LTL formula $\varphi$ (denoted $tr \models \varphi$) if and only if $(tr,0) \models \varphi$.
Moreover, the value function $V$ is extended to LTL formula $\varphi$ and trace $tr$ as $V(\varphi,tr) = 1 \leftrightarrow (tr,0)\models \varphi$.

\subsection{LTL Learning Problem with Misclassification}
Learning the LTL formula from demonstrations is essential for capturing desired temporal requirements of complex systems without demanding manual and skilled effort.

The LTL learning problem considers a given finite sample of traces $T = (P, N)$ divided into a set of positive demonstrations $P$ and negative demonstrations $N$.
The set \( P \) contains positive traces representing desired behaviors aligned with our intended expectations, while the set \( N \) consists of negative traces illustrating behaviors that the system should avoid \cite{neider2018learning}.

Although the data samples $T$ are divided into positive and negative traces, in practice, some demonstrations may be incorrectly labeled due to noise.
To quantify these misclassifications, 
a loss function that measures the fraction of traces incorrectly categorized is introduced \cite{gaglione2022learning}.
\begin{equation}
\ell(T, \varphi) = \frac{1}{|T|}\left( \sum_{tr \in P} |V(\varphi, tr)-1| + \sum_{tr \in N} |V(\varphi, tr)| \right)
\end{equation}
Based on this function, the learning problem is defined as follows:

\begin{problem}[from \cite{gaglione2022learning}]
\label{prob:ltl_neider}
Given a set of traces \( T = (P, N) \) and a misclassification threshold \( \kappa \in [0,1] \), find a \emph{minimal} LTL formula \( \varphi \) such that:

\begin{equation}
\ell(T, \varphi) \leq \kappa.
\end{equation}
The objective is to find the smallest possible formula, in terms of operators and propositional variables, that describes the system's behavior with at most $\kappa$ misclassification loss. 
Shorter formulas are generally easier to interpret and offer clearer insights. Although a larger formula could achieve zero loss, it would be harder to understand and thus less useful.
\end{problem}
\vspace{-0.55em}
\section{Problem Statement}
\label{sec:problem}



When dealing with real-world data, uncertainties and perturbations might affect the data collection process. 
For instance, uncertainties stem from sensor noise, sensor faults, communication delays, or unexpected interactions that influence the observed data. 
As a result, the collected data does not solely reflect the intended behavior of the system. 
In this section, we introduce the problem of learning LTL formulas from data demonstrations with uncertainty.
\vspace{-0.3em}
\subsection{Trace Uncertainty}
To capture and evaluate the uncertainty in the data, we define the concept of \emph{trace uncertainty}.
We assume that each \emph{observed} trace may contain deviations from the true underlying system behavior.
Thus, the information about the \emph{correct trace} is uncertain among the \emph{observed trace}. 
From the observed trace, we want to define all possible \emph{trace estimates}, i.e., the set of possible correct traces that can generate the observed trace.

In this work, we quantify the possible differences in the traces using the Hamming distance \cite{hamming1950error}.
Given two traces of equal length, the Hamming distance measures the number of replacements needed to transform one trace into the second one. 
Formally, the Hamming distance is a function  
$d_H : \Sigma^n \times \Sigma^n \rightarrow \mathbb{N}$, defined as:
$$d_H(tr_1, tr_2) = \sum_{i=0}^{n} |tr_1[i] \triangle tr_2[i]|$$ where $\Sigma = 2^{\AP}$, $n\in \mathbb{N}$ is the length of the string, and $\triangle$ is the symmetric difference set operator\footnote{$A\triangle B = (A\setminus B) \cup (B\setminus A)$}.
Thus, $d_H(t_1,t_2)$ quantifies how many atomic propositions differ between traces $t_1$ and $t_2$.

\begin{example}
Let the set of atomic propositions be $\AP = \{a,b\}$. 
The Hamming distance between traces $t_1 = \{a\}\{b\}$ and $t_2 = \{\}\{b\}$ is equal to $1$.
\end{example}

Based on the Hamming distance, we need to connect the uncertainty between an observed trace and the possible trace estimates. 
Herein, we assume that a known \emph{uncertainty bound} is given.
This bound represents that the uncertainty in traces is bounded, e.g., bounded noise. 

\begin{definition}
Given a trace $t$ and a bound $k$, the set of trace estimates is defined as $\mathcal{C}_{tr}^k = \{tr'\in \Sigma^{|tr|}\mid d_H(tr,tr')\leq k\}$.
The set of trace estimates $\mathcal{C}_{tr}^k$ is called the \emph{trace estimate group} for trace $tr$.
\end{definition}

Intuitively, the set $\mathcal{C}_t^k$ contains all possible traces that have Hamming distance at most $k$ from trace $t$.
In our learning context, trace $t$ is the observed trace with uncertainty bounded by $k$.
To account for the uncertainty, the set $\mathcal{C}_t^k$ represents a set of trace estimates that can represent the correct behavior. 
These candidates, referred to as \emph{trace estimates}, represent possible corrections to the observed data under uncertainty assumptions.
We apply this bound because the recorded trace might not match what actually occurred, as it can be distorted by measurement errors, partial observability, communication delays, or corrupted data. 



\begin{example}
Let $\AP = \{a, b\}$, trace $t = \{a\}\{b\}$, and uncertainty bound $k= 1$ be given.
The set of trace estimates \( C^k_{t} \) includes all traces that differ from \( t \) in at most one position by one atomic proposition.
\[
C^1_{t} = \left\{
    \{a\}\{b\},\ 
    \{\}\{b\},\ 
    \{a, b\}\{b\},\ 
    \{a\}\{a, b\},\ 
    \{a\}\{\}
\right\}
\]
Trace in \( C^1_{t} \) represents an 
estimate of the observed trace \( t \).
\end{example}

Since we consider trace estimates from the set of observed trace samples $T =(P,N)$, we extend $T$ to include the set of estimates.
Let $T_{est} = (P_{est},N_{est})$ be the extended set of traces that includes all possible positive and negative traces within a bound $k$.
Formally, $P_{est} = \cup_{tr\in P} \mathcal{C}_{tr}^k$ includes every possible positive trace estimate for traces in $P$ and bound $k$.
Similarly, $N_{est}= \cup_{tr\in N} \mathcal{C}_{tr}^k$ every possible negative trace estimate.
With an abuse of notation, we write $T_{est}$ to represent the union of all trace estimates, $P_{est}\cup N_{est}$.

\vspace{-0.5em}
\subsection{Learning Robust LTL Formulas}



As previously discussed, Problem~\ref{prob:ltl_neider} cannot handle uncertainty in the traces since it \emph{assumes correct traces}.
Although label misclassifications are incorporated in Problem~\ref{prob:ltl_neider}, traces are assumed to be correctly sampled.
Therefore, we need a new learning problem that considers the uncertainty bounds created by the set of possible trace estimates.

Within each set of trace estimates $C^k_{tr}$, we assume that a ``correct" trace belongs to this set, including the observed trace $t$.
In this manner, our learning problem considers at least one trace within each set of trace estimates during the learning process.
Formally, from the set of all positive trace estimates $P_{est}$, the learning algorithm considers a subset $P'\subseteq P_{est}$ that: (i) is consistent with the learned formula $\varphi$ -- for any $tr\in P'$ then $tr\models \varphi$, (ii) selects at least one trace estimate from each uncertainty group -- for any $tr\in P$ then $\mathcal{C}^k_{tr}\cap P'\neq \emptyset$.
We can define similar constraints for negative traces, where the only difference is the consistency with $\varphi$, i.e., for any $tr\in N'$ then $tr\not\models \varphi$.
Therefore, our learning problem does not assume that the observed trace is exactly correct, but rather that it is one of several estimates within its uncertainty set. 

To account for misclassification errors in our learning problem, we generalize the idea of misclassification to a new concept called trace group misclassification. 
Given a trace $tr\in P$, a set of trace estimates $P'\subseteq P_{est}$, and an uncertainty bound $k$, group misclassification considers that at least one trace estimate $tr'$ from the uncertainty group $\mathcal{C}^k_{tr}$ (i) is consistent with learned formula $\varphi$ ($tr'\models \varphi$) and (ii) is selected ($tr'\in P'$).
To encode this misclassification loss, we define the group misclassification function $l^P_g$ for positive traces $tr\in P$, a set of trace estimates $P'\subseteq P_{est}$, bound $k$, and formula $\varphi$:
\begin{equation}
\ell_g^P(tr, P' , k, \varphi) = 
\begin{cases}
1 & \text{if } \Gamma=\emptyset \text{ or } \exists tr'\in \Gamma.\ V(tr', \varphi) = 0 \\
0 & \text{otherwise}
\end{cases}
\end{equation}
where $\Gamma = \mathcal{C}^k_{tr}\cap P'$.
Again, the definition of the loss function $l_g^N$ for negative traces is similar to $l_g^P$, with the consistency condition $\exists tr'\in \Gamma.\ V(tr', \varphi) = 1$.
By combining these two loss functions, we define the loss function \( \ell_g(T', k, \varphi) \), which quantifies the total misclassification for trace estimate set $T' = (P', N')$ with $P'\subseteq P_{est}$ and $N'\subseteq N_{est}$.
\begin{equation}
\ell_g(T,T', k, \varphi) = \frac{1}{|T|}\left( \sum_{t \in P} \ell_g^P(t,P', k, \varphi) + \sum_{t \in N} \ell_g^N(t,N', k, \varphi) \right)
\label{eq:robust_loss}
\end{equation}
Function $l_g$ aggregates the individual loss terms \( \ell_g^P \) and \( \ell_g^N \) over all traces and normalizes by the total observed traces $T$.

Based on the misclassification functions, the robust learning problem can now be stated as follows:
\begin{problem}[Robust Learning LTL with Trace Uncertainty] \label{prob:robust}
Given a misclassification threshold \( \kappa \in [0, 1] \), the objective is to find a minimal LTL formula \( \varphi \) such that:
\begin{equation}
\ell_g(T, T', k, \varphi) \le \kappa.
\end{equation}
\end{problem}

The goal is to learn the minimal LTL formula that fits demonstrations under bounded trace uncertainty while keeping group misclassification within an acceptable tolerance $\kappa$.


\section{Learning Robust LTL Formulas Algorithm}
\label{sec:solution}
To address the robust learning problem under trace uncertainty, we build on prior work that infers LTL formulas using MaxSAT optimization \cite{neider2018learning,gaglione2022maxsat}.
Our main contribution is to extend these approaches to handle uncertainty in the demonstration traces.
We introduce a propositional formula $\Phi$, constructed from the LTL conditions, the demonstrations $T$, the maximum formula size $n$, and an uncertainty bound $k$.
Together with cost assignments, $\Phi$ encodes all constraints of Problem~\ref{prob:robust}.
The formula is satisfiable if and only if there exists an LTL formula $\varphi^*$ that solves Problem~\ref{prob:robust}.

\begin{figure}[h!]
    \centering
    \includegraphics[width=0.7\linewidth]{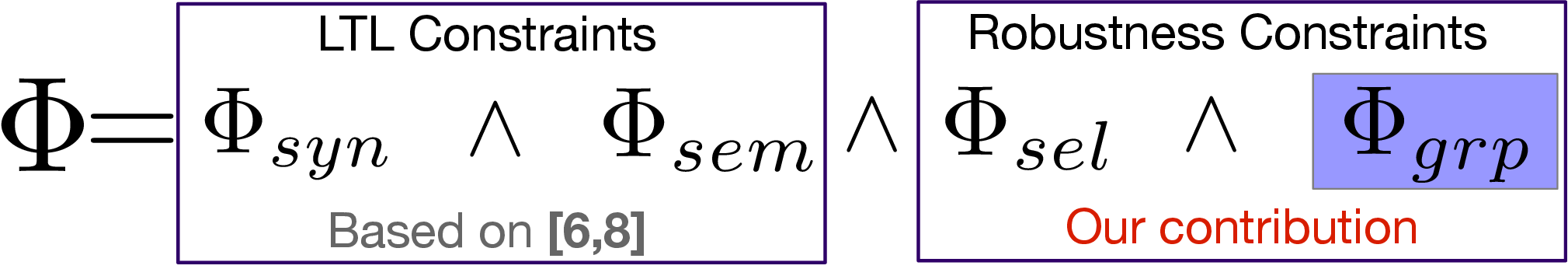}
    \caption{Propositional Formula $\Phi$ overview construction. The formula $\Phi_{grp}$ highlighted in blue is the only soft constraint in MaxSMT.}
    \label{fig:prop-constraints}
    \vspace{-1em}
\end{figure}
Figure~\ref{fig:prop-constraints} illustrates how the propositional formula $\Phi$ is constructed from two categories of constraints: LTL constraints and robustness constraints.
The LTL constraints ensure that the learned formula follows the syntax of LTL ($\Phi_{syn}$) and that the traces satisfy its semantics ($\Phi_{sem}$).
The robustness constraints determine which trace estimates are selected during learning ($\Phi_{sel}$) and enforce satisfaction across groups of trace estimates ($\Phi_{grp}$).
In the next sections, we first formally define each of these subformulas and specify their associated cost assignments.
Next, we present additional constraints on the LTL properties to allow efficient generation of LTL formulas.
Lastly, we present optimization constraints to ensure the solution of Problem~\ref{prob:robust}.

\vspace{-0.45em}
\subsection{LTL Constraints}

\noindent
\textbf{Syntax Constraints: }
Following the work of \cite{neider2018learning,gaglione2022maxsat}, each candidate LTL formula is represented by a generic syntax tree with a fixed number of nodes $n$.
Each tree node $i\in \{1,\dots, n\}$ corresponds to an atomic proposition $\AP$ or a logical/temporal operator $\{\neg,\vee,\wedge,U,\dots\}$. 
Structural Boolean constraints ensure the correct construction of the tree.
These constraints are captured using Boolean variables:  $x_{i,a}$ indicates that node $i$ is labeled with symbol $a \in \AP \cup \{\neg,\vee,\wedge,U,\dots\}$, and $l_{i,j}, r_{i,j}$ indicate left and right children.
For example, $x_{3,\vee}$ and $r_{3,2}$ indicate that node $3$ is the $\vee$ operator and has node $2$ as its right children.
The tree is enforced to satisfy the LTL syntax constraints and tree constraints, such as one root node, leaf nodes are in $\AP$, unary operators only have one child, etc.
See \cite{neider2018learning} for more details.




\noindent
\textbf{Semantic Constraints: }
As in \cite{neider2018learning}, to verify if a trace $tr$ satisfies the LTL formula encoded in $\Phi_{syn}$, we need to introduce a Boolean formula $\Phi_{tr}^n$.
Intuitively, this Boolean formula is satisfied if and only if trace $tr$ satisfies the LTL formula encoded in $\Phi_{syn}$.
To construct $\Phi_{tr}^n$, we define Boolean variables $ y_{i,t}^{tr}$ to capture the satisfaction of the subformula rooted at node $i$ at time $t$, i.e., $y_{i,t}^{tr}$ is satisfied if and only if the suffix of $tr$ starting at $t$ satisfies the subformula rooted at node $i$. 
To ensure that these variables have this desired meaning, every LTL operator has its Boolean formula.
For instance, the semantic rule for negation is expressed as:
\begin{equation}
(x_{i,\neg} \land l_{i,j}) \rightarrow \bigwedge_{0 \leq t < |t'|} \left( y_{i,t}^{t'} \leftrightarrow \neg y_{j,t}^{t'} \right).
\end{equation}
This ensures that if node $i$ is labeled with $\neg$ and its child is node $j$,
then the satisfaction of node  $i$ at time $t$ is the negation of the satisfaction at node $j$. 
For the complete set of semantic constraints, see \cite{neider2018learning,gaglione2022maxsat}.
The construction of $\Phi_{sem}$ is defined by the conjunction of all traces, $\Phi_{sem} = \bigwedge_{tr \in T_{est}}\Phi_{tr}^n$.

\vspace{-0.45em}
\subsection{Robustness constraints}


\noindent
\textbf{Trace Selection Constraints: }
When defining the semantic constraints, we introduced the variables $y^{tr}_{i,t}$ that correspond to the values $V(\varphi_i, tr, t)$, i.e., if trace $tr$ at step $t$ satisfies LTL formula $\varphi_i$ rooted at node $i$.
Therefore, the variable $y^{tr}_{n,0}$ represents if $tr\models \varphi$.
In \cite{neider2018learning}, the semantic constraints include a satisfaction condition for positive traces $tr\in P$, $\Phi^n_{tr} \wedge y^{tr}_{n,0}$, and for negative traces  $tr\in N$, $\Phi^n_{tr} \wedge\neg y^{tr}_{n,0}$.
In this manner, these constraints ensure that trace $tr$ satisfies (does not satisfy) the formula encoded by the syntax tree, i.e., $tr\models \varphi$ ($tr\not \models \varphi$).
These constraints are considered as soft constraints in \cite{gaglione2022maxsat}, which allows a fraction of them to be violated.
However, in our setting, traces consist of trace estimates rather than ``exact" traces as in \cite{neider2018learning,gaglione2022maxsat}.

Trace $tr\in T$ represents an observed trace, while traces in $tr'\in C_{tr}^k$ are the possible trace estimates that could have generated observation $tr$.
Different than the previous works, we cannot enforce that all positive (negative) trace estimates satisfy (do not satisfy) the learned formula. 
First, trace estimates might belong to both sets of positive and negative trace estimates.
Second, enforcing satisfaction at the level of each trace estimate would constrain the solution by treating uncertain estimates as ground truth.

To overcome this problem, we introduce a set of binary \emph{selection variables} $z _{tr}$ for $tr \in T_{est}$.
Intuitively, $z_{tr}$ encodes whether a trace estimate $tr$ is taken into account in the learning.
For instance, given a positive trace estimate $tr\in P_{est}$, if $z_{tr} = 1$, then $tr$ needs to satisfy the learn formula of the syntax tree, $tr\models \varphi$.
In other words, if we consider a trace estimate to infer an LTL formula, this trace must be consistent with the learned formula.
To ensure that the variables $z_{tr}$ reflect this desired interpretation, we impose the following constraints:
\begin{align}
&z_{tr} \rightarrow y_{n,0}^{tr}, \ tr\in P_{est}\label{eq:pos_sel_constraint}\\
&z_{tr} \rightarrow \neg y_{n,0}^{tr},\ tr\in N_{est}\label{eq:neg_sel_constraint}
\end{align}

The propositional formula in Eq.~(\ref{eq:pos_sel_constraint}) implements the condition that a positive trace estimate needs to satisfy, $y_{n,0}^{tr}$, the learned formula if $z_{tr}$ is set to $1$.
On the other hand, Eq.~(\ref{eq:neg_sel_constraint}) enforces that a negative trace estimate does not satisfy the learned formula when $z_{tr}=1$.
In this manner, we do not enforce that every trace estimate is consistent with the learn formula, but only the traces selected, i.e., $z_{tr}=1$.
To enforce these constraints on all trace estimates, we define the selection $\Phi_{sel}$ as:
\begin{equation}
\Phi_{sel} = \bigwedge_{tr\in P_{est}}(z_{tr} \rightarrow y_{n,0}^{tr})\ \wedge \bigwedge_{tr\in N_{est}}(z_{tr} \rightarrow \neg y_{n,0}^{tr})
\end{equation}

\noindent
\textbf{Trace Group Constraints: }
Although the trace selection constraint enables the choice of which trace estimates are considered during learning, a trivial solution is to set $z_{tr} = 0$ for all estimates.
While posing Problem~\ref{prob:robust}, we assume that within each set of trace estimates $\mathcal{C}_{tr}^k$ a ``correct'' trace belongs to this set.
Thus, we impose that at least one trace estimate from each uncertainty group must be selected in the learning of the LTL formula.


\begin{equation}
\Phi_{grp} = \bigwedge_{tr \in P \cup N} \left( \bigvee_{tr' \in \mathcal{C}_{tr}^k} z_{tr'} \right)
\end{equation}

By combining the four ``core" constraints as in Fig.~\ref{fig:prop-constraints}, we have the core propositional formula to solve Problem~\ref{prob:robust}.
Next, to complete our solution, we introduce additional constraints that allow more efficient LTL learning.
\vspace{-0.45em}
\subsection{Additional Syntactic Constraints}
In addition to the core constraints, we introduce further syntactic constraints on the use of LTL operators and atomic propositions in the generated formula, similar to \cite{zhang2025constrained}.
These additional constraints come directly from the user, who can introduce them as prior knowledge.
Herein, we consider that the user introduces constraints regarding a critical subset of atomic propositions $\AP_{\text{critical}}\subseteq \AP$.
Formally, for each atomic proposition $p \in \mathcal{P}_{\text{critical}}$, there must exist some node $i$ in the formula such that $x_{i,p}$ is true.
\begin{equation}
\Phi_{usr} = \bigwedge_{p \in \AP_{\text{critical}}} \left( \bigvee_{i \leq n} x_{i,p} \right)
\end{equation}


\subsection{Optimization Objective}



\begin{table*}[ht]
    \centering
    \caption{Summary of Comparison Results for HVAC system}
    \label{tab:combined_table}
    \footnotesize
    \begin{tabularx}{\linewidth}{@{}
    c
    >{\centering\arraybackslash}p{0.7cm}
    c c c c
    X c
    X c
    X c@{}}
        \toprule
        & \multicolumn{4}{c}{\textbf{Uncertainty Window}} 
        & \multicolumn{2}{c}{\textbf{Flie}} 
        & \multicolumn{2}{c}{\textbf{Modified Flie}} 
        & \multicolumn{2}{c}{\textbf{Robust Learning [Ours]}} \\
        \cmidrule(lr){2-5} \cmidrule(lr){6-7} \cmidrule(lr){8-9} \cmidrule(lr){10-11}
        & \textbf{Start} & \textbf{End} & \textbf{Var} & \textbf{k}
        & \textbf{Formula} & \textbf{Time (s)} 
        & \textbf{Formula} & \textbf{Time (s)} 
        & \textbf{Formula} & \textbf{Time (s)} \\
        \midrule
        & 3 & 4 & AC & 1
        & $\lnot \text{hot} \; \mathsf{U} \; \mathsf{X}\,\text{hot}$ & 4.4 
        & $\mathsf{G}\,(\text{AC} \rightarrow (\mathsf{F}\,\text{Occ}))$ & 2.3 
        & $\mathsf{G}(\text{AC} \rightarrow \text{Occ})$ & 16.7 \\[6pt]

        & 1 & 4 & hot & 1
        & $(\mathsf{F}\,\text{AC} \land \lnot \text{AC})$ & 8.2 
        & $\mathsf{G}(\text{AC} \rightarrow \text{Occ})$ & 0.86 
        & $\mathsf{G}(\text{AC} \rightarrow \text{Occ})$ & 27.2 \\[6pt]

        & 2 & 3 & Occ & 1
        & $ \mathsf{X}\,\text{hot}$ & 0.09
        & $\text{AC} \rightarrow \mathsf{G}\,\text{Occ}$ & 0.8 
        & $\mathsf{G}(\text{AC} \rightarrow \text{Occ})$ & 6.5 \\[6pt]

        & 3 & 4 & Occ & 1
        & $(\mathsf{F}\,\text{AC} \land \lnot \text{AC})$ & 4.4
        & $(\mathsf{F}\,\text{AC}) \;\mathsf{U}\; \text{Occ}$ & 0.84 
        & 
$
\big( \, (\, \mathsf{F}\,\text{Occ} \;\mathsf{U}\; \text{AC} \,) \;\rightarrow\; \text{AC} \, \big)
\;\;\rightarrow\;\;
\mathbf{G}\,\text{AC}
$ & 73.1 \\[3pt]
        \bottomrule
    \end{tabularx}  
    \vspace{-2em}
\end{table*}

So far, we have encoded all constraints as propositional formulas. 
If this encoding is fed directly to a SAT solver, the solution will necessarily satisfy all constraints. 
While feasible, such a solution does not address the optimization in Problem~\ref{prob:robust}. 
To incorporate the group misclassification loss function $l_g$, we define it based on the Boolean variables $x, y, z$. 
Similar to \cite{gaglione2022maxsat}, we rely on Max-SMT \cite{sebastiani2015optimization}. 
However, unlike \cite{gaglione2022maxsat}, where optimization is expressed via partial weighted Max-SAT (i.e., weights on clauses in CNF), our loss function is formulated as a Pseudo-Boolean optimization objective.

All constraints other than $\Phi_{grp}$ are hard constraints: $w(\Phi_{syn})=w(\Phi_{sem})= w(\Phi_{sel}) = w(\Phi_{usr}) = \infty$.
The constraint $\Phi_{grp}$ is defined as a soft constraint.
To optimize the group constraint, we define a weight for each sample trace $tr\in T$:
\begin{align}
\mathrm{w}(tr) &=\;\frac{1}{|T|}  
\left(\prod_{tr' \in C_{tr}^k} (1-z_{tr'}) \right)
\end{align}
Intuitively, the weight $w(tr)$ is equal to $\frac{1}{|T|}$ if all traces in the uncertainty group $\mathcal{C}_{tr}^k$ are \emph{not} selected, i.e., $z_{tr'} = 0$ for all $tr'\in \mathcal{C}_{tr}^k$.
Note that if a trace $tr'\in \mathcal{C}_{tr}^k$ is selected, $z_{tr'} =1$, then the selection constraint $\Phi_{sel}$ enforces its semantic correctness, i.e., $tr'$ satisfies the formula if $tr'\in P_{est}$ or $tr'$ does not satisfy the formula if $tr'\in N_{est}$.
Formally, our optimization function is defined as:
\begin{equation}\label{eq:opt-fun}
wl = \sum_{tr\in T} w(tr)
\end{equation}
In this manner, the Pseudo-Boolean Optimization is to minimize $wl$, which encodes the loss function $l_g$.
\section{Evaluation}
\label{sec:experiments}

In this section, we compare our robust learning algorithm to the Max-SAT-based (Flie) learning algorithm by \cite{gaglione2022maxsat}.
Since Flie does not allow user constraints $\Phi_{usr}$, we also modified Flie, called Modified Flie, to allow these types of constraints.
We implemented our algorithm in Python using the Microsoft Z3 SMT-solver \cite{demoura2008z3}.
We compare the formulas learned by Flie and Modified-Flie against our learning method when uncertainty is presented in the samples.
For our evaluation purposes, we first instantiate a ground-truth formula specification.
Based on this formula, we generate a set of positive and negative traces.
We then introduce uncertainty into the traces by randomly modifying them with respect to location in the trace, affected atomic proposition, and maximum uncertainty bound.
All methods are evaluated using these uncertain traces to learn formulas, while the ground-truth specification is used solely for comparison.

\subsection{HVAC Control System Specification}
First, we illustrate our proposed methods using an HVAC (Heating, Ventilation, and Air Conditioning) control system.
For this system, we define the atomic propositions set $\AP = \{hot,\ Occ,\ AC\}$.
The AP ${hot}$ indicates whether the room temperature exceeds a predefined threshold.
$Occ$ specifies if the room is occupied, while $AC$ determines if the AC is on.
In this example, we want to evaluate the methods for different uncertainties affecting the sample traces, e.g., location, variable, and bound.

The ground-truth formula specification for the HVAC system is $\mathbf{G}((hot \wedge Occ) \rightarrow \mathbf{X}\,AC)$, i.e., always when the room is occupied and hot, then the $AC$ needs to be on in the next step.
In Table~\ref{tab:combined_table}, the uncertainty window column summarizes the uncertainty in the traces: the start and end positions mark the segment of the trace where uncertainty occurs, while the variable column specifies which variable is affected.
The column $k$ indicates the maximum uncertainty bound, i.e., the maximum Hamming distance.
Regarding the learning, we fix the maximum misclassification error of $\kappa=0.25$.

%



\begin{figure}[h!]
    \centering
    \includegraphics[width=0.7\linewidth]{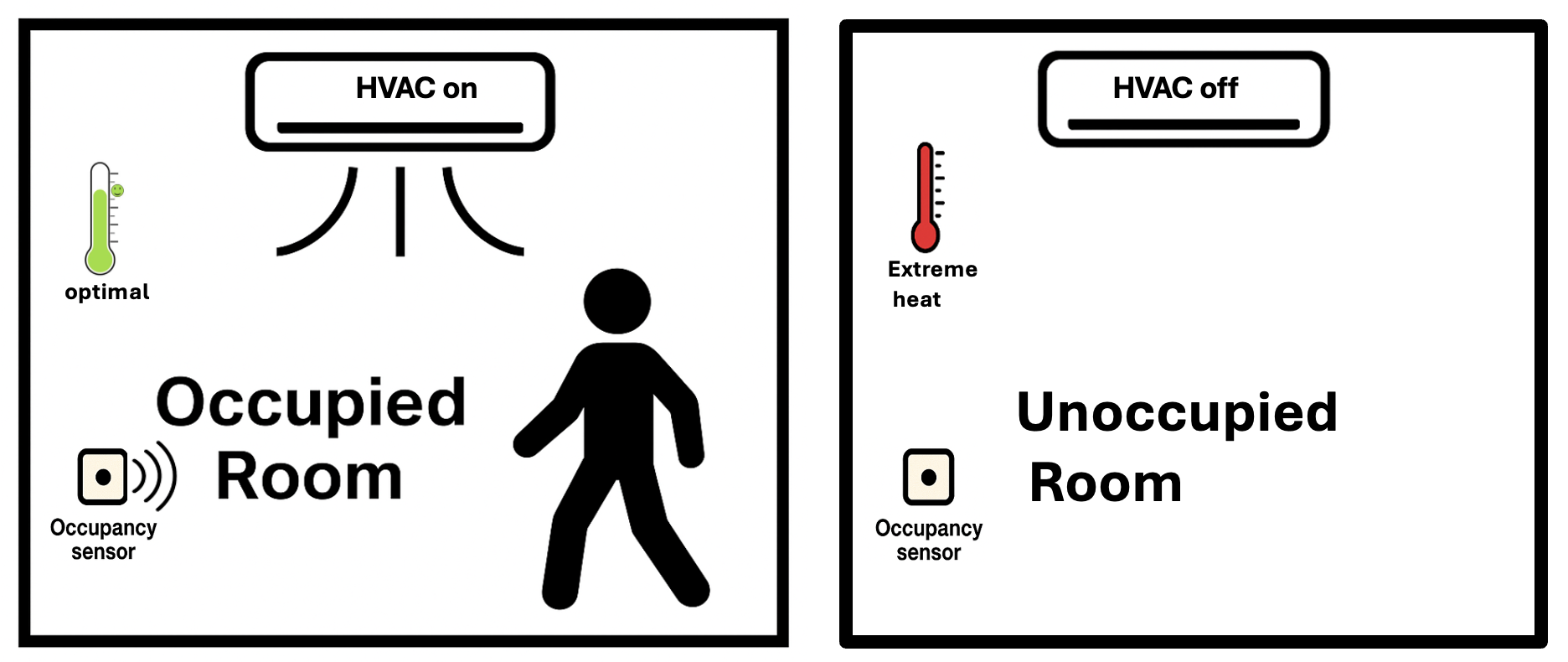}
    \caption{HVAC System Operation with and without Occupancy}
    \label{fig:HVAC}
    \vspace{-1.5em}
\end{figure}

Table~\ref{tab:combined_table} also summarizes the comparison among the three algorithms.
It provides the learned formula based on the uncertain traces, as well as the time to learn the formula.
We observe that Flie learned simple formulas without finding the temporal relationship among atomic propositions, e.g., $\neg hot\ \mathsf{U}\ \mathsf{X} hot$.
This result is expected since Flie is not designed to handle trace uncertainty.

The Modified Flie runs Flie with the additional $\Phi_{usr} = \{AC,\ Occ\}$ constraint, which significantly improves the learned formula.
In fact, in three out of the four cases, Modified Flie recovers some temporal relationships present in the ground-truth specification, i.e., $\mathsf{G}(AC\rightarrow \mathsf{F}Occ)$ and $\mathsf{G}(AC\rightarrow Occ)$.
However, the learned formulas invert the relation: instead of $Occ$ and temperature triggering the $AC$ as in the ground truth, the $AC$ implies occupancy, and omitting the temperature condition and the precise temporal step.

Similarly, our robust learning method partially recovers temporal relationships in all four cases, e.g., $\mathsf{G}(AC\rightarrow Occ)$.
Like Modified Flie, the learned formula flips the causal direction compared to the ground truth.
However, our robust learning approach prioritizes immediate temporal dependencies between atomic propositions and consistently generates specifications that are more aligned with the intended system behavior.
While modified FLIE may preserve certain aspects of the temporal structure, our approach emphasizes critical causal patterns, such as ensuring that occupancy is detected before AC activation, rather than allowing potentially inefficient behaviors like preemptive actuation.
Note that our method took more time to learn the formulas.
This behavior is expected, since Pseudo-Boolean optimization extends Max-SAT by incorporating weighted linear constraints, making the problem computationally more challenging.

\begin{table*}[t]
    \centering
    \setlength{\tabcolsep}{5pt}
    \renewcommand{\arraystretch}{1.1}
    \caption{Comparison of robust learning with the state-of-the-art method under different uncertainty windows.}
    \label{tab:combined_table1}
    \begin{tabular}{l c c c c l c l c l c}
    \hline
    \multicolumn{1}{c}{\textbf{Ground-truth Formula}} &
    \multicolumn{4}{c}{\textbf{Uncertainty Window}} &
    \multicolumn{2}{c}{\textbf{Flie}} &
    \multicolumn{2}{c}{\textbf{Modified Flie}} &
    \multicolumn{2}{c}{\textbf{Robust Learning [Ours]}} \\
    \cline{2-5} \cline{6-7} \cline{8-9} \cline{10-11}
     & \textbf{Start} & \textbf{End} & \textbf{Var} & \textbf{k}
     & \textbf{Formula} & \textbf{Time (s)}
     & \textbf{Formula} & \textbf{Time (s)}
     & \textbf{Formula} & \textbf{Time (s)} \\
    \hline
    $\mathbf{G}\!\left(x_{1} \rightarrow \mathbf{G}\,x_{0}\right)$
    & 1 & 1 & $x_1$ & 1
    & $\neg x_{1}$ & 0.16
    & $\left(x_{1} \rightarrow \mathbf{G}\,x_{0}\right)$ & 1.3
    & $\mathbf{G}\!\left(x_{1} \rightarrow x_{0}\right)$ & 4.3
    \\
    $\mathbf{F}\,x_{1} \rightarrow (\neg x_{0}\,\mathsf{U}\,x_{1})$
    & 3 & 3 & $x_1$ &1
    & $x_{1}$ & 0.09
    & $(x_{0} \rightarrow x_{1})\ \mathsf{U}\ x_{1}$ & 6
    & $\neg x_{0}\ \mathsf{U}\ x_{1}$ & 23.5
    \\
    $\mathbf{G}\,\neg x_{0}$
    & 4 & 4 & $x_0$ &1
    & $\mathbf{G}\,x_{0}$ & 1.8
    & $\neg x_{0}$ & 2
    & $\neg x_{0}$ & 7.1
    \\
    $(\mathbf{G}\,\neg x_{0}) \lor \mathbf{F}\!\big(x_{0}\land \mathbf{F}\,x_{1}\big)$
    & 2 & 2 & $x_1$&1
    & $\mathbf{X}\,x_{0}$ & 0.13
    & $\big(x_{1} \rightarrow \mathbf{X}\,x_{0}\big)$ & 1.7
    & $
\big( x_{1} \;\rightarrow\; x_{0} \big) \;\mathsf{U}\; x_{0}
$ & 5.2
    \\
    $\mathbf{F}\,x_{0}$
    & 3 & 3 & $x_0$ &1
    & $x_{0}$ & 0.05
    & $x_{0}$ & 0.05
    & $x_{0}$ & 0.05
    \\
    $\mathbf{F}\,x_{1} \rightarrow (x_{0}\ \mathsf{U}\ x_{1})$
    & 3 & 6 & $x_1$ &4
    & $x_{1}$ & 0.03
    & $x_{0}\ \mathsf{U}\ x_{1}$ & 1.04
    & $x_{0}\ \mathsf{U}\ x_{1}$ & 30.5
    \\
    \hline
    \end{tabular}
    \vspace{-1.5em}
\end{table*}
\vspace{-0.45em}
\subsection{Common LTL$_f$ patterns}

To further evaluate the performance of our approach, we conducted experiments on additional synthetic datasets.
Here, we compare the methods using different LTL$_f$ formulas that can be found in practice, i.e., compare the methods for different formulas generating traces with uncertainty.

Table~\ref{tab:combined_table1} summarizes the results of our experiments.
We ran the learning methods based on 6 different LTL$_f$ formulas with different uncertainty on them and $\kappa$ values varying from 0.25 to 0.14.
As expected, the Flie method did not perform well, often learning formulas that discard temporal information. 
The modified Flie performs better by recovering parts of the temporal relationships present in the ground-truth formulas in some cases.
On the other hand, our robust learning method was able to recover parts of the temporal relationships present in the ground-truth formulas in five out of the six cases.
For instance, in the first row, Modified Flie learns 
\(x_{1} \rightarrow \mathbf{G}x_{0}\), which captures a one-step implication, whereas the ground-truth formula \(\mathbf{G}(x_{1} \rightarrow \mathbf{G}x_{0})\) expresses a persistent condition across all timesteps.
Our robust method, in contrast, preserves the global structure.

Overall, across both benchmark sets, the robust approach tends to yield formulas that are logically closer to the ground-truth specification
This advantage comes from the ability of our method to incorporate uncertainty before learning the LTL formulas.
In contrast, existing methods assume that the provided traces are correct, which often leads them to learn spurious temporal relationships or to overlook temporal dependencies when uncertainty is present.
Although our results are still preliminary, these findings indicate that the robust approach offers a more principled and promising direction compared to the current state of the art.

\vspace{-0.55em}
\section{CONCLUSIONS}
\label{sec:conclusion}

This paper presented a robust framework for learning Linear Temporal Logic (LTL) specifications from system traces that may be noisy, incomplete, or affected by measurement errors. We addressed uncertainty in trace using the Hamming distance and introduced group-based constraints to ensure that at least one trace from each uncertainty group contributes to the learning process. 
This approach enables the inference of LTL formulas that are minimal and capture essential system behaviors, even when the input data has uncertainty. 
The framework also incorporates expert knowledge as constraints to guide learning toward more meaningful and relevant specifications.
Overall, our method bridges the gap between theoretical LTL learning and the practical challenges faced in real-world systems. 
As future work, we aim to develop and apply formal distance metrics to quantitatively compare the learned specifications against known ground-truth formulas, providing a more precise evaluation of learning accuracy and behavioral similarity.





\bibliographystyle{IEEEtran}
\bibliography{romulo_bib} 

\end{document}